# Extending planning knowledge using ontologies for goal opportunities

Mohannad Babli, Universitat Politècnica de València, Department of Computer Systems and Computation, Valencia, Spain, mobab@dsic.upv.es

Eva Onaindia, Universitat Politècnica de València, Department of Computer Systems and Computation, Valencia, Spain, onaindia@dsic.upv.es

Eliseo Marzal, Universitat Politècnica de València, Department of Computer Systems and Computation, Valencia, Spain, emarzal@dsic.upv.es

**Abstract.** Approaches to goal-directed behaviour including online planning and opportunistic planning tackle a change in the environment by generating alternative goals to avoid failures or seize opportunities. However, current approaches only address unanticipated changes related to objects or object types already defined in the planning task that is being solved. This article describes a domain-independent approach that advances the state of the art by extending the knowledge of a planning task with relevant objects of new types. The approach draws upon the use of ontologies, semantic measures, and ontology alignment to accommodate new acquired data that trigger the formulation of goal opportunities inducing a better-valued plan.

## Introduction

Planning research has been mostly devoted to offline planning with some incursions in online plan-repair to address failures during the plan execution. Whilst online planning has demonstrated its usefulness to handle plan failures, unanticipated events that bring about an opportunity for the task at hand has been rarely studied. Goal-directed behaviour (GDB) is a hallmark of intelligence widely used for high levels of autonomy when the environment is dynamic, partially observable, and open to new data (Vattam et al. 2013). In GDB, the agent monitors the execution of the plan in the environment and it is capable of formulating alternative goals on the fly (Cox 2007; Dannenhauer and Muñoz-Avila 2013; Klenk, Molineaux, and Aha 2013). One limitation of most of the current GDB approaches is that goals are formulated on the basis of objects that already exist in the agent model. An exception to this can be found in (Cashmore et al. 2017), an approach to *opportunistic planning* which allows the agent to generate new goals involving objects that are not present in its current model. Nevertheless, the new object must be of one of the predefined classes (types) in the agent model.

The motivation of this work is to overcome the general lack of research in GDB towards the formulation of goal opportunities that stem from objects or object classes that are unknown to the agent. Specifically, given a planning task and a plan (sequence of actions) that solves the task, the process initiates with the execution of such a plan. While executing the plan, events from the environment are received. External events are classified into three categories: events that confirm the correct execution of the plan actions; events that bring about a failure in the plan execution; or events that may induce a new goal opportunity in the context of the planning task and the plan. This paper puts the focus on the latter and proposes an approach to handle context-aware open planning tasks.

Our contribution is a domain-independent approach that advances the state of the art by extending the knowledge of a planning task with relevant objects extracted from a collection of ontologies that describe features of interest for the specific domain. Our approach draws upon the richness and expressivity of standard ontology representations, semantic measures and ontology alignment for accommodating the new acquired objects into the planning task specification. These new objects may subsequently trigger the formulation of a goal that induces a better-valued plan.

## Background

Consider a tourism scenario in which a tourist wishes to make a one-day tour to visit several points of interest (POIs) in a city. The scenario is formulated as a planning task including a specified set of POIs of different categories (place types), the opening and closing times of POIs, the walking time between locations and a list of potentially visitable POIs to the tourist (recommendable POIs or task goals). The planner solves this task and returns a plan which includes a total of four visits: two POIs of type *religious site*, an *emblematic architectural building* and a visit to a POI of type *aquarium*. During the plan execution, the tourist receives a cellphone notification and learns about a small exhibition on Picasso's paintings that has opened nearby. This external information includes a new object type, *art exhibition*, not formerly considered in the planning task, which may represent an opportunity to the tourist if the goal of visiting *Picasso's exhibition* can be aligned within the modelling of the planning task and triggers a plan compliant with the current goals that results in a better tourism experience to the tourist.

A planning task is defined as $\Phi = \langle \mathcal{D}, \mathcal{I} \rangle$, where $\mathcal{D}$ is the domain of the task (e.g. tourism) and $\mathcal{I}$ is a particular problem instance (e.g. a one-day tour for a specified tourist). The elements that define the domain are $\mathcal{D} = \langle \mathcal{T}, \mathcal{V}, \mathcal{A} \rangle$: $\mathcal{T}$ is the set of object types (e.g. types of POIs); $\mathcal{V}$ is the set of boolean variables of the form $(p\ o_1 \cdots o_n)$, where $p$ is a predicate symbol and arguments $\{o_i\}_{i=1}^{n}$ are of types included in $\mathcal{T}$ (e.g. `(at ?tourist ?location)`); and $\mathcal{A}$ is the available action schemas (e.g. `(visit ?tourist ?POI)`). On the other hand, an instance is described by $\mathcal{I} = \langle \mathcal{O}, \mathcal{S}, \mathcal{G} \rangle$ where $\mathcal{O}$ is the set of objects (e.g. POIs of the city); $\mathcal{S}$ is a full assignment of values to variables in $\mathcal{V}$ that represent the current state of the problem ($|\mathcal{S}| = |\mathcal{V}|$ and initial values of $\mathcal{V}$ denote the initial state of the task); and $\mathcal{G}$ is a partial assignment of values to variables of $\mathcal{V}$ that represent the goals to be accomplished by the plan computed by the planning solver (e.g. set the variable `(visited JohnPerry cathedral)` to true). The planner receives $\Phi$ as input and outputs a plan $\pi = \langle a_1 \cdots a_n \rangle$ composed of a sequence of ground actions (e.g. `(move JohnPerry hotelGarden cathedral) (visit JohnPerry cathedral)` $\cdots$).

Plan monitoring consists in observing the state that results from executing the plan actions in the environment and checking whether the *observed state* $\mathcal{S}$ matches the *expected state*. This operation creates a *discrepancy set* as the difference between the two sets, which will comprise instantiated variables that are found in the observed state but not in the expected state and vice versa. After discarding the variables of the discrepancy set that represent a failure, the remaining variables denote a potentially achievable goal opportunity. More specifically, the discrepancy set will contain instantiated variables of the form $(p\ o_1 \cdots o_n)$ where $\exists\ o \in \{o_i\}/o \notin \mathcal{O}$. Note that there is no distinction between an object $o$ that belongs to one of the existing types in $\Phi$ or an object of a new type. Both cases are treated similarly from the planning point of view and the only difference lies in the use of the ontology mechanism. An example of discrepancy set in the tourism domain is { `(open StCatherineChapel)`, `(open PicassoExhibition)` }, where the first object `StCatherineChapel` belongs to the existing type *religious site* and the object `PicassoExhibition` belongs to the new discovered object type *art exhibition*.

## Outline of the approach

Figure 1 sketches our approach which works in four stages:

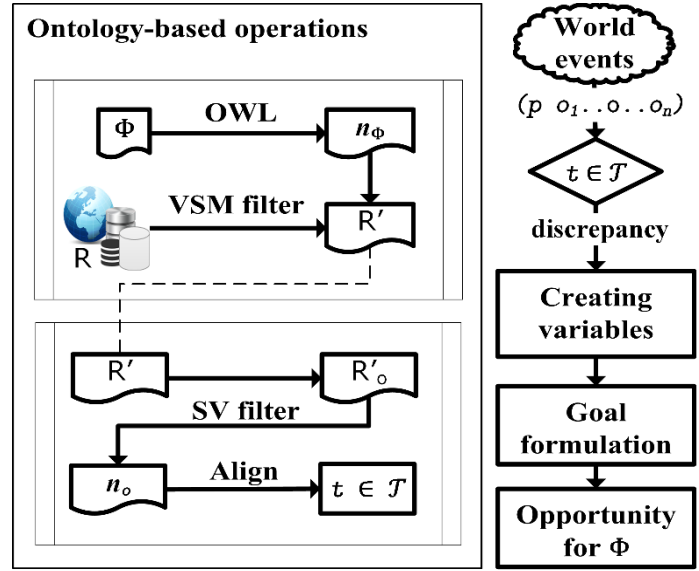

Figure 1: The ontology-based Goal formulation model

Stage 1: Identification of similar ontologies. First, we create an OWL ontological representation of the types $\mathcal{T}$ and objects $\mathcal{O}$ of $\Phi$, called $n_\Phi$, that will be the base ontological representation for the rest of stages. Second, we retrieve a set $R$ of remote ontologies from on-line repositories (such as Mondeca Tourism Ontology that includes important concepts of the tourism domain e.g., locations and accommodations, and concepts that describe leisure activities and geographic data. Subsequently, we apply a vector space distance similarity measure (VSM) to $R$ and we obtain the set $R'$, which contains the most similar ontologies to $n_\Phi$.

Stage 2: Positioning a new object. When the information of a new object $o$ ($o \notin \mathcal{O}$) is received in the form of a variable $(p\ o_1 \cdots o_n)$, the system tries to find $o$ in $R'$ and creates $R'_o$ as the set of ontologies out of $R'$ that contain $o$. The ontology of $R'_o$ that most accurately models the semantic knowledge of the application domain $\mathcal{D}$ is selected using the semantic variance measure (we will refer to this ontology as $n_o$). Next, the system retrieves the type $t$ of $o$ using $n_o$. In case that $t \in \mathcal{T}$, (literally the same type) we simply add $o$ to $\mathcal{O}$. Otherwise, the system attempts to position $t$ in $n_\Phi$ via a semantic alignment with a neighbourhood constraint between $n_\Phi$ and $n_o$. If the alignment is successful, $t$ is either identified as an existing type in $\mathcal{T}$ and we simply add $o$ to $\mathcal{O}$, or $t$ is found to be a new type, then we add it to $\mathcal{T}$ and $o$ is added to $\mathcal{O}$.

Stage 3: Creating the new variables. If the new object $o$ is successfully added in $\Phi$, the next step is to instantiate the required planning variables $\mathcal{V}$, besides $(p\ o_1 \cdots o_n)$, that describe $o$. The system automatically identifies the information required for integrating $o$ in $\Phi$, requests this information from Open Data platforms (e.g. geographic coordinates of a new location), and adds it to $\mathcal{S}$.

Stage 4: Goal formulation. If the type $t$ of object $o$ is a type or a sibling of a type that is involved in a goal $g \in \mathcal{G}$, then we formulate $x$ candidate new goals that involve the newly received object $o$, where $x$ depends on the possible permutations of objects in the goal predicate; $x = 1$ if the goal has only $o$ as a parameter such as $g = (q\ o)$ (e.g. $g =$ (visited PicassoExhibition)).

The remaining sections of the paper detail the OWL ontological representation of the types, the identification of similar ontologies, the selection of the ontology with the highest semantic insight, the alignment with a neighbourhood constraint, the identification of an opportunity, elaborated cases of study to demonstrate the behaviour of our system with a representative example, and finally we conclude.

**Ontology-based operations**

In this section we detail the tasks required to extend the planning task knowledge to include new objects.

*OWL Ontology representation*

OWL has been the World Wide Web Consortium recommendation since 2004 (Patel-Schneider, Hayes, and Horrocks 2014). In this section, we explain the OWL ontology representation of $n_\Phi$. We utilised OWL API which is an open source Java API and reference implementation for creating, manipulating, and serialising OWL Ontologies (Horridge and Bechhofer 2011). Throughout this section we use snapshots from the GUI of Protégé to show visual explanations of $n_\Phi$.

The planning ontological representation consists of a set $\mathcal{C}$ of concepts (OWL classes) that are used to represent $\mathcal{T}$; a set of OWL annotation properties used to describe $\mathcal{C}$, and the set of individuals that represent $\mathcal{O}$. A class $c \in \mathcal{C}$ can have one or many annotation properties. The symbol :: is used to refer to a sub class, for instance, $c_i :: c_j$ means $c_j$ is a subclass of $c_i$.

On the other hand, the Planning Domain Description Language (PDDL) (Edelkamp and Hoffmann 2004) offers the ability to express a type structure for the objects in a domain, typing the parameters that appear in predicates and actions. Furthermore, types can be expressed as forming a particular type hierarchy. For each type in $\mathcal{T}$, an OWL class is created in $n_\Phi$ abiding the exact hierarchy. The left part of Figure 2 shows the types $\mathcal{T}$ of the *tourism* domain specified in PDDL and the right part of the figure shows the corresponding $\mathcal{C}$ in $n_\Phi$. For instance, the type aquarium is represented as $c_{\text{aquarium}} :: c_{\text{attraction}}$. We can observe the types of the planning task are arranged in a reasonable hierarchy and that the OWL representation follows truthfully this hierarchy. Although using real names of types that convey a semantic meaning and having the types arranged in a reasonable hierarchy do not affect the ontological representation in anyway, these aspects are important to find ontologies that represent similar domains and for positioning $t$ in $\mathcal{T}$. Nonetheless, the dependency of using real names and a significant type hierarchy in the system does not affect its domain-independent nature.

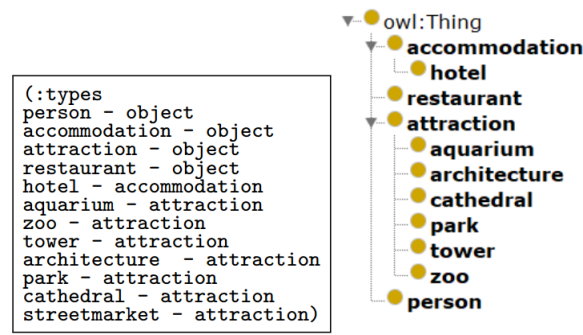

Figure 2: Representation of PDDL types

The definition of a particular problem instance $\mathcal{I}$ includes a declaration of a set of objects $\mathcal{O}$ (an example is visualized in Figure 3). For each $o \in \mathcal{O}$ of type $t \in \mathcal{T}$, an OWL individual o is created in $n_\Phi$ of the class $c_t$; e.g., Lonja is an object of the PDDL type architecture, the system creates Lonja as an individual of the class $c_{\text{architecture}}$ as shown in the lower part of Figure 3.

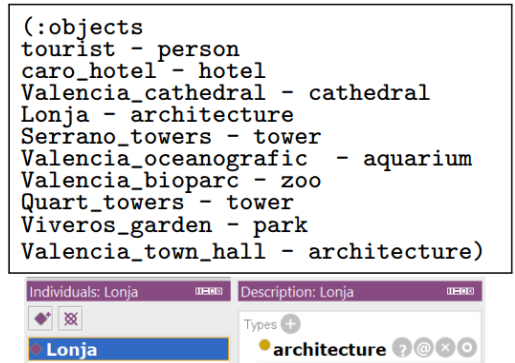

Figure 3: PDDL objects of a particular tourism planning task

*Identifying similar ontologies*

In ontology engineering, it is useful to know quickly if two ontologies are close or remote before deciding to match them (David and Euzenat 2008). It is not always the case that a golden reference ontology is available for any applications domain, and even if it exists the agent should be able to find it autonomously. In this step, we measure the distance between $n_\Phi$ and the ontologies of R to filter out the unrelated ones and obtain R′. Furthermore, we need to tackle the natural complication that different people could model the same application domain using different terms, or even in different languages; e.g., ontologies A and B in Figure 4 model the tourism domain using different terminology, and ontologies C and D model a product delivery domain also using different terminology.

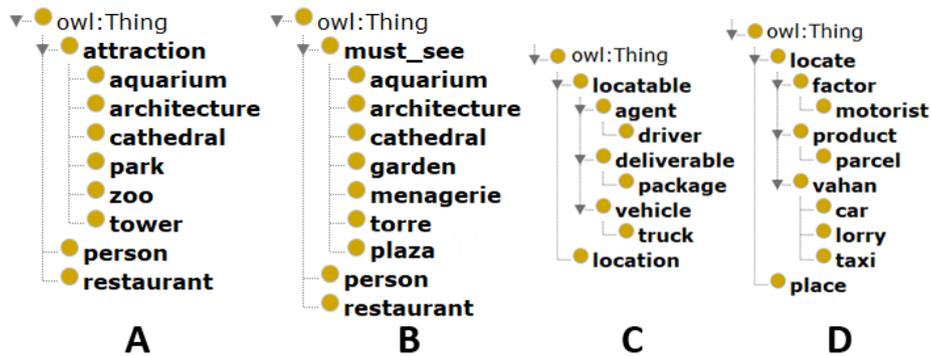

Figure 4: Using different terms when modelling ontologies

For that purpose, we use ConceptNet as a standard mean to describe the classes of the ontologies. ConceptNet (Speer, Chin, and Havasi 2017) is a knowledge graph that connects words and phrases of natural language using labelled edges. Its knowledge is obtained from various sources that combines expert-created resources, crowd-sourcing, and games with a purpose. ConceptNet utilises a closed class of 36 selected relations such as isA, usedFor, hasProperty, etc., with the aim of representing relationships independently of the language or the source of the terms it connects. Therefore, we augment the classes of $n_\Phi$ and of the ontologies of R with the relations and classes brought from ConceptNet as OWL annotations. As a result, even if the names of the classes are different, classes that refer to the same concept will have annotations in common and will be found similar when measuring semantic distances or when performing the alignment. Figure 5 shows a small portion of the twenty one annotations attached to the class $c_{\text{attraction}}$.

Annotations: attraction
Annotations
rdfs:label  [type: xsd:string]
antonym repulsion
rdfs:label  [type: xsd:string]
atLocation disneyland
rdfs:label  [type: xsd:string]
causesDesire flirt

Figure 5: Annotations sample assigned to Class attraction

For measuring the distance between ontologies we decided to look at the ontologies as a bag of terms; consequently, ontology distance measures based on the Vector Space Model (VSM) are applicable (VSM filter in Figure 1). Measuring similarity in the VSM using cosine index with TF (weighted term frequency) has proven to obtain good results compared to other distance measures and it is computed largely faster, but it is not much robust to lexical alterations (David and Euzenat 2008). However, lexical alterations do not impact our approach thanks to classes are augmented with OWL annotations coming from ConceptNet as a standard means to describe concepts. Thereby, the lexical information in each term of the ontology comes not only from the local name of the term but also from annotations imported from ConceptNet relations and classes. In order to compute the distance we used *OntoSim*, an independent Java API that allows to compute similarities between ontologies developed with *OWL API* and provides a variety of distance measures. At this stage, the ontologies R′ (shown in Figure 1) with the highest similarity with respect to $n_\Phi$ are obtained.

*Selecting the ontology with the highest semantic insight*

The system calculates R′o, the set of ontologies from R′ that contain the individual o. Then, we raise the famous question: which ontology, $n_o$, amongst the ontologies of R′o, which all cover a certain domain of knowledge and model the same concepts, is the best for the specific task Φ? For this purpose we decided to use semantic variance (SV filter in Figure 1) that is introduced in (Sánchez et al. 2015). SV is an intuitive and inherently semantic measure to evaluate the accuracy of ontologies. Unlike ad hoc methods, SV is a mathematically coherent extension of the standard numerical variance to measure the semantic dispersion of the taxonomic structure of ontologies. The SV formal definition is introduced by Sánchez (2015); given an ontology $n$, which models in a taxonomic way a set of concepts $\mathcal{C}$ in $n$, the SV of $n$ is computed as the average of the squared semantic distance between each concept $c_i \in \mathcal{C}$ and the taxonomic root node of $n$. If we denote by $|\mathcal{C}|$ the cardinality of $\mathcal{C}$ excluding root, the mathematical expression of SV of $n$ is:

$$SV = \frac{\sum_{c_i \in \mathcal{C}} d(c_i, root)}{|\mathcal{C}|}$$

Where $d(c_i, c_j)$ is the semantic distance between two concepts $c_i$ and $c_j$ calculated as a function of the number of their non-common taxonomic ancestors divided (for normalisation) by their total number of ancestors:

$$d(c_i, c_j) = \log_2 \left( 1 + \frac{|A(c_i) \cup A(c_j)| - |A(c_i) \cap A(c_j)|}{|A(c_i) \cup A(c_j)|} \right)$$

Where $A(c_i)$ is the set of taxonomic ancestors of concept $c_i$ in $n$, including itself. The semantic distance measure $d$ aggregates features in a logarithmic way, which better correlates with the non-linear nature of semantic evidences, and more importantly, variance does not depend on the cardinality of the ontology.

We calculate the SV for each ontology in R'o, and select $n_o$, the one with the highest SV. Next, the system retrieves the class $c_t$ of the individual o in $n_o$, in case $c_t$ exists in $n_\Phi$, it means $t \in \mathcal{T}$ (literally the same term) and we simply add $o$ to $\mathcal{O}$. Otherwise, $t \notin \mathcal{T}$ and then the system attempts to position the class $c_t$ (which corresponds to the type $t$ in $n_\Phi$ as explained in the following section.

*Alignment with neighbourhood constraint*

The next step is to determine where to position the class $c_t$ within the hierarchy of concepts $\mathcal{C}$ in $n_\Phi$. We perform an alignment between, on one hand, $\mathcal{C}$ in $n_\Phi$, and on the other hand, the particular part of the taxonomic branch of $n_o$ that includes $c_t$, the parent class $c_{\text{parent}(t)}$, and the siblings $\mathcal{C}_{\text{siblings}(t)}$.

For the purpose of the alignment we used CIDER-CL introduced in (Shvaiko et al. 2013), a schema-based ontology alignment system that compares two ontologies by analysing their similarity at different levels of their ontological context, and then combines the similarities by means of artificial neural networks. CIDER-CL also makes use of VSM, more specifically SoftTFIDF, a hybrid string similarity measure that combines TF-IDF with an edit-based similarity distance to support a higher degree of variation between the terms. This makes CIDER-CL very suitable in our approach due to the fact that we augmented the classes using ConceptNet. If $c_t$ matches one of the classes in $n_\Phi$, it means the two classes refer to the same type $t$ albeit they are syntactically different. In this case, $t$ is found to be semantically equivalent to an existing type of $n_\Phi$ and thus we simply add $o$ to $\mathcal{O}$. Otherwise, we use a neighbourhood constraint, since domain-independent constraints convey general knowledge about the interaction between related nodes and perhaps the most used such constraint is the neighbourhood constraint, as suggested in (Doan et al. 2003) where "*two nodes match if nodes in their neighbourhood also match*". If $c_{\text{parent}(t)}$ in $n_o$ matches $c_x$ in $n_\Phi$, then we establish $c_x :: c_t$ in $n_\Phi$. On the other hand, if no match was achieved with the parent, the neighbourhood constraint procedure matches $\mathcal{C}$ of $n_\Phi$ with $\mathcal{C}_{\text{siblings}(t)}$ of $n_o$, if the percentage of matching siblings exceeds a specified threshold, and these matched classes are found to be under a common parent in $n_\Phi$, then we list $c_t$ as a subclass of that superclass in $n_\Phi$. If the alignment was successful, we add $t$ to $\mathcal{T}$ and $o$ to $\mathcal{O}$. For instance, consider the ontologies A and B in Figure 4, $\mathcal{C}$ in A represents the classes of A, and $\mathcal{C}$ in B represent the classes of B, for instance, let the newly received variable be (open Virgen plaza); the new object $o$ is Virgen plaza, and its type $t$ is found to be plaza $\notin \mathcal{T}$, the alignment finds that $c_{\text{parent}(\text{plaza})}$ is $c_{\text{must\_see}}$ in B and it matches $c_{\text{attraction}}$ in A; therefore, the system asserts $c_{\text{attraction}} :: c_{\text{plaza}}$ in A, and creates the individual Virgen plaza of the class $c_{\text{plaza}}$ in A; correspondingly (new object of a new type), a new entry plaza - attraction is added to $\mathcal{T}$, and $Virgen\ plaza$ is added to $\mathcal{O}$.

## Opportunity identification

Once the $x$ candidate new goals are formulated as explained in Stage 4 of the overview of our approach, the system generates $\Phi' = \Phi'_1, \cdots, \Phi'_x$ (modified versions of $\Phi$), where the added information includes $o$, $t$, the information of $o$, the discrepancy proposition, $\mathcal{G}' = g'_i \cup \mathcal{G}$, and the new current state $\mathcal{S}$. We use a planner to solve each $\Phi'_i \in \Phi'$ to know which $g'_i$ can be considered an opportunity to $\Phi$ in the context of $\pi$. $g'_i$ is considered an opportunity goal for $\Phi$, when the planner is able to generate a plan $\pi'_i$ to solve $\Phi'_i$ that includes the new goal plus the original set of goals, taking into consideration the metric of the planning problem. For simulating the execution, and monitoring the environment we reutilised the simulation system in (Babli et al. 2016) that takes the planning task $\Phi$ and the plan $\pi$ as input and encodes them into a timeline as a collection of chronologically ordered timed events encapsulating the changes to be expected in the subsequent states. The monitoring process, on one hand, simulates 1) external events and adds it to the timeline, and on the other hand, 2) the execution of each timed event, checking that conditions are successfully satisfied and the effects happen when they should, thus validating and updating, respectively, the states of the world.

## Cases of study

The aim of this section is to show the behaviour of our system with a representative example. Consider a *Tourism* Φ, in which a tourist wishes to make a one-day tour in the city of Valencia. Initially, the system retrieves a set of recommended places according to the user profile and a set of restaurants. $\mathcal{T}$ is shown in Figure 2, $\mathcal{O}$ is shown in Figure 3, and $\mathcal{V}$, $\mathcal{A}$, and $\mathcal{G}$ are shown in Figure 6, respectively.

```
(:predicates
  (be ?tourist - person ?loc -
    (either accommodation attraction restaurant))
  (visited ?tourist - person ?loc - attraction)
  (eaten ?tourist - person)
  (open ?loc - (either attraction restaurant))
  (time_for_eat ?tourist - person)
  (free_table ?loc - restaurant)
  (active ?tourist - person))

(:durative-action move ...)
(:durative-action visit ...)
(:durative-action eat ...)

(:goals (and
  (visited tourist Cathedral)
  (visited tourist Lonja)
  (visited tourist Serrano_towers)
  (visited tourist Quart_towers)
  (visited tourist Viveros_garden)
  (eaten tourist)
  (be tourist Caro_hotel)))
```

Figure 6: $\mathcal{V}$, $\mathcal{A}$, and $\mathcal{G}$ in Tourism Φ

The tourist location is at `Caro hotel`. The tourist is active between 10:00 and 23:00 and wishes to eat between 13:00 and 19:00. The opening/closing times of the POIs and the restaurants, the recommended durations of POIs visits, and the duration of movement between locations are imported from Open Data platforms and included in $\mathcal{S}$. The tour to solve Φ (PLAN1 shown in the top left snapshot of Figure 7) is calculated by the planner. The visits included in PLAN1 are marked with red location pins in the snapshot. The tour starts from the origin location of the tourist, i.e., the hotel in which the user is staying at (green location pin), and includes visits to five attractions (red pins) and one stop at a restaurant (orange pin).

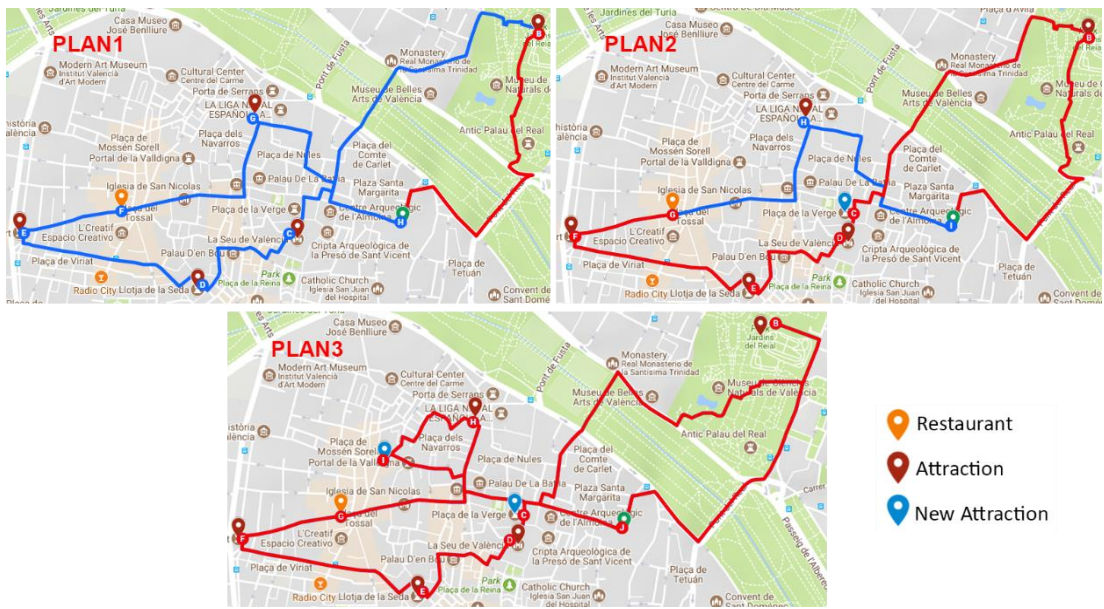

Figure 7: The three simulated plans. Pins in PLAN1: (A/H) Caro hotel, (B) Viveros Garden, (C) Cathedral, (D) Lonja, (E) Quart towers, (F) El Celler del Tossal (RESTAURANT), and (G) Serrano towers. Pins in PLAN2: (A/I), B are the same as PLAN1, (C) Virgen plaza, (D) Cathedral, (E) Lonja, (F) Quart towers, (G) El Celler del Tossal (RESTAURANT), and (H) Serrano towers. Pins in PLAN3: (A/J), B, C, D, E, F, G, H are the same as PLAN2, and (I) Jimmy Glass Jazz bar.

The system creates $n_\Phi$ (shown in Figure 8 A), then it accesses several ontologies available in online-repositories R′ = {B, C, D, E, F} (shown in Figure 8). The ontologies are augmented using ConceptNet. VSM distance is calculated between A, and each ontology in R′, the distances are respectively: 0, 0, 0.64, 0.79, and 0.78. R′ = {D, E, F} is recognized as the set of most similar ontologies.

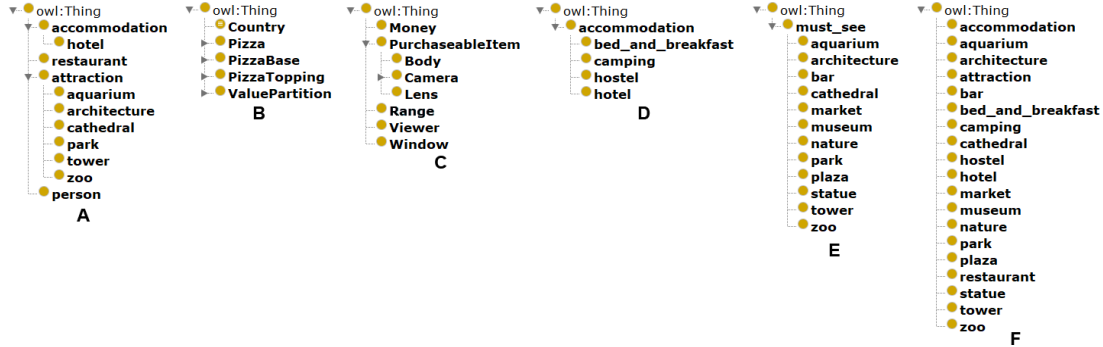

Figure 8: Example of several remote ontologies

The simulator starts PLAN1 execution simulation. Let us assume that after visiting the first attraction `Viveros Garden` (after the red line in PLAN1, Figure 7), a new information is received (`open Virgen plaza`). The system tries to find Virgen plaza in R′ and creates R′$_o$ = {E, F}, the set of ontologies out of R′ that contain `Virgen plaza`. The SV distance is measured to find whether E or F describes more accurately the semantics of the application domain, the values are respectively 0.25 and 0.16; therefore, $n_o$ = E. Next, the system retrieves the type of `Virgen plaza` from E that is plaza $\notin \mathcal{T}$ and attempts to position $c_{plaza}$ in A by aligning A and E. The system finds that $c_{parent(plaza)}$ in E is $c_{must\_see}$ and it matches the class $c_{attraction}$ in A; therefore $c_{attraction} :: c_{plaza}$ inside A. An individual Virgen plaza of the class $c_{plaza}$ is created in A. Correspondingly, a new entry plaza - attraction is added to $\mathcal{T}$, and Virgen plaza is added to $\mathcal{O}$. The information required for integrating `Virgen plaza` in $\Phi$ is automatically identified and requested from Open Data platforms; e.g. movement durations between the new location and the existing locations. This information is added to $\mathcal{S}$. Since plaza is a sibling of a type that is involved in a goal $g \in \mathcal{G}$ thus, the system formulates a new $g'$ = (`visited tourist Virgen plaza`), and $\mathcal{G}' = g' \cup \mathcal{G}$. Finally, the system updates $\mathcal{S}$ with the current state at the time the new information was received. The planner is called to generate a new plan (PLAN2 shown in Figure 7); allowing the tourist to visit the original set of POIs plus the new POI. The simulation continues. Let us assume that after the tourist has eaten in the restaurant *el celler del tossal*, a new information is received (`openJimmy Glass Jazz bar`), similarly, the system deals with the new information and extends the knowledge of the planning task, a new plan is obtained (PLAN3 shown in Figure 7), and the simulation continues. At the end of the day, the tourist ends up in visiting seven attractions instead of five attraction, eating in the restaurant, and returning to the hotel.

## Conclusion

Context and context awareness are crucial for any intelligent agent that operates in a dynamic environment. To develop context-aware ambient intelligence planning service, suitable context models and reasoning approaches are necessary. In this paper we have presented a domain-independent approach that may be considered as a context model and a first step towards a context aware ambient intelligent planning service. Our approach bolsters an autonomous agent with the capability of extending its planning task to accommodate new information on the fly; to learn information about the planning task and to introduce relating information such as new objects whether of existing types or more importantly new types during the execution of the initial plan that solves the original planning task, that in turn may trigger the formulation of new goals and produce new plans to achieve the new goals in addition to the original set of goals. On the other hand, for future work we intend to focus on the goal reasoning approaches and to endow the system with the ability to perform goal reasoning instead of delegating that task to a planner.

## Acknowledgment

This work is supported by the Spanish MINECO project TIN2017-88476-C2-1-R. The authors wish to thank Cassens J. and Wegener, R. for helpful discussions.